\documentclass{article}
\usepackage{ijcai26}

\usepackage{times}
\usepackage{soul}
\usepackage{url}
\usepackage[hidelinks]{hyperref}
\usepackage[utf8]{inputenc}
\usepackage[small]{caption}
\usepackage{graphicx}
\usepackage{amsmath}
\usepackage{amsthm}
\usepackage{booktabs}
\usepackage{algorithm}
\usepackage{algorithmic}
\usepackage[switch]{lineno}
\usepackage{multirow}

\title{Six Layers Less: Encoder Pruning for Whisper with Label-Free Recovery}

\author{Rasmus Aagaard$^{1,2}$ \and Nicki Skafte Detlefsen$^1$ \affiliations $^1$Technical University of Denmark\\ $^2$Laerdal Medical\\ \emails \{roraa, nsde\}@dtu.dk, }

\begin{document}

\maketitle

\begin{abstract}
    Pruning large pre-trained transformer-based ASR models such as OpenAI's Whisper has seen great adoption, as pruning the decoder led to significant end-to-end transcription speedups. For instance, the {\tt whisper-large-v3-turbo} variant reduced the decoder from 32 to 4 layers, while Distill-Whisper similarly reduced the decoder to only 2 layers. Although some attention has been put towards reducing the size of the encoder, no approach has seen wide adoption. This could be due to the need for custom inference implementations to take advantage of the compressed model. We present an approach that ranks encoder layers by the leave-one-layer-out change in Word Error Rate (WER). The six layers that cause the least change are removed, corresponding to $18.5\%$ of the encoder stack. The pruned model requires no custom inference code as it is simply a more shallow encoder with fewer layers. We further distill using unlabeled monolingual speech data to recover performance degradation caused by the zero-shot layer pruning. Mean WER across four languages increases to $20.1\%$ after distillation, compared to $21.9\%$ zero-shot, going from a baseline of $18.2\%$. We release all of our code\footnote{\url{https://github.com/rasgaard/whisper-encoder-layer-prune}} and the pruned model\footnote{\url{https://huggingface.co/rasgaard/whisper-large-v3-turbo-encoder-pruned}}.
\end{abstract}

\section{Introduction}
Voice interfaces relying on Automatic speech recognition (ASR) systems are increasingly implemented in applications such as medical transcriptions and real-time captioning. OpenAI's Whisper \cite{radford2023robust}, an encoder-decoder transformer \cite{Vaswani2017NEURIPS_Attention_is_All}, is highly regarded as a performant model and has seen wide adoption since its release. Several successful attempts have been made to compress Whisper \cite{Gandhi2023ARXIV_Distil_Whisper_Robust,Orhon2025ARXIV_WhisperKit_On_device,Kamahori2025EMNLP_LiteASR_Efficient_Automatic}, indicating that the original model contains redundancies. Finding redundancies and compressing deep neural networks is an important area of research for better accessibility and deployment options, especially for resource constrained devices.

We explore this idea of identifying and exploiting redundancies in Whisper further by pruning entire layers of the often overlooked encoder. Observing that removal of many of the individual layers in the encoder causes only a small change in the Word Error Rate (WER), we hypothesize that many of the layers can be dropped together without any major degradation to transcription performance.

\begin{figure}[tb]
    \centering
    \includegraphics[width=1\linewidth]{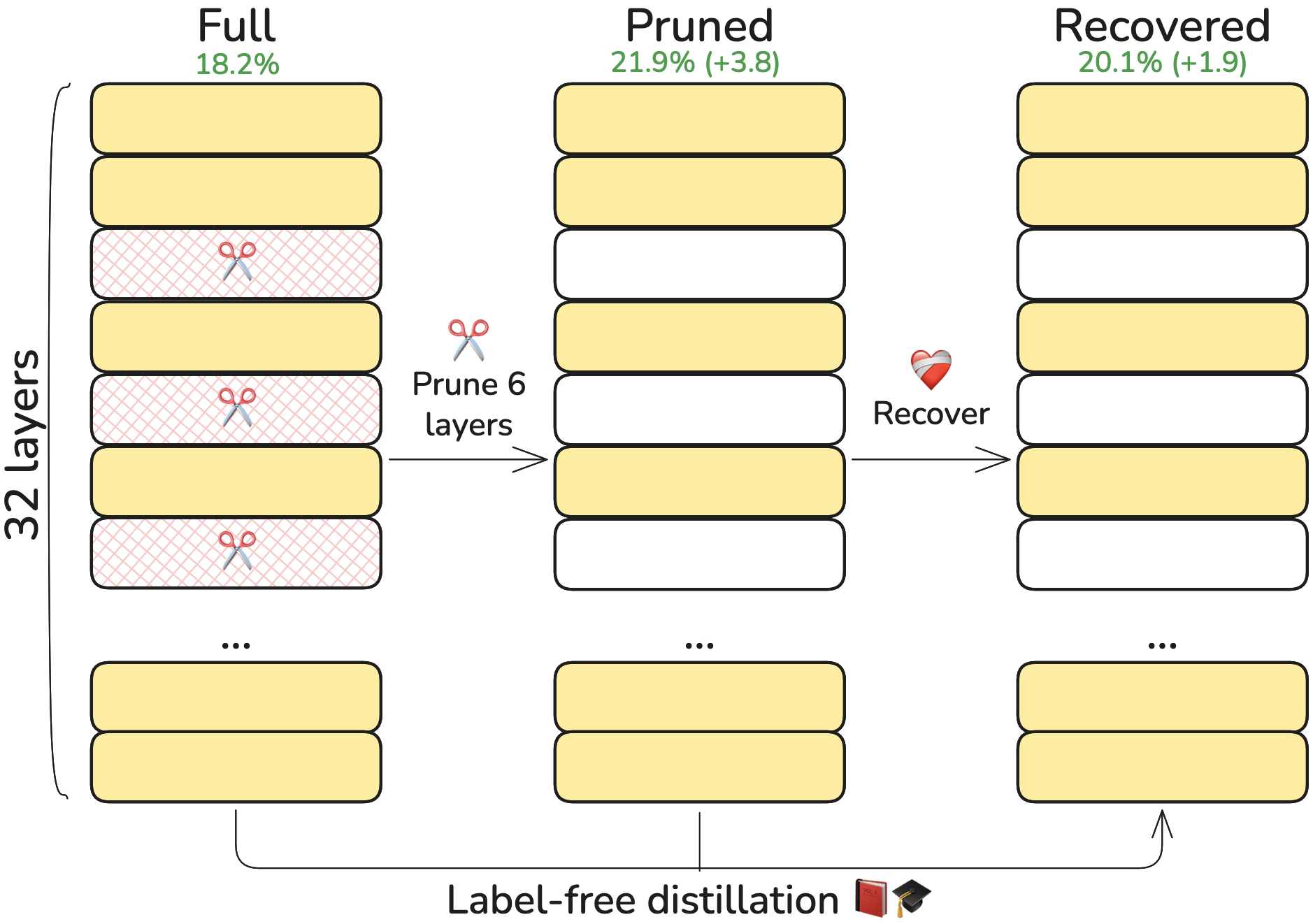}
    \caption{Transcription performance throughout the process of pruning the layers in {\tt whisper-large-v3-turbo}'s encoder, going from the full encoder (left), zero-shot pruned encoder (middle) and the encoder recovered through modest label-free distillation (right).}
    \label{fig:sketch}
\end{figure}

We find that it is possible to prune six layers from Whisper's encoder with minor degradations to performance. In addition, we close the gap even further using Knowledge Distillation \cite{hinton2015distilling}, viewing the original encoder as the teacher and the pruned encoder as the student. This teaches the student to produce hidden states similar to the teacher through Mean Squared Error (MSE) loss.   

In summary, our work makes the following contributions.
\begin{enumerate}
    \item We introduce a simple layer pruning approach to reducing {\tt whisper-large-v3-turbo}'s encoder stack by 18.5$\%$ that can effectively be deployed through existing inference libraries without architectural changes.
    \item We show that multilingual performance lost from zero-shot layer pruning can largely be recovered through label-free knowledge distillation at a modest training and data budget. 
    \item We show that data-driven selection of layers for pruning is essential by comparing the performance degradation to random layer selections.
\end{enumerate}

\section{Methods}

Inspired by work on the ineffectiveness of certain layers in Large Language Models (LLMs) \cite{Gromov2025ICLR_The_Unreasonable_Ineffectiveness} we view the Whisper encoder under a similar lens. We observe that the residual connection during the forward pass carries the majority of the signal throughout the encoder, making the hidden states highly similar throughout the layers.

Recent work suggests that it is insufficient to measure layer importance using the traditional cosine similarity method of computing the similarity between the input and output states of a given layer \cite{Hinostroza2026ICLR_Rethinking_Layer_Relevance}. Following this approach, we rank the 32 layers in the Whisper encoder by their $\Delta$WER, measuring the WER on the FLEURS \cite{conneau2023fleurs} test set with both the full encoder and the 32 leave-one-layer-out encoders. Considering multilingual performance we compute the mean $\Delta$WER on four languages: Danish, English, German and French. 

\begin{figure}[tb]
    \centering
    \includegraphics[width=\linewidth]{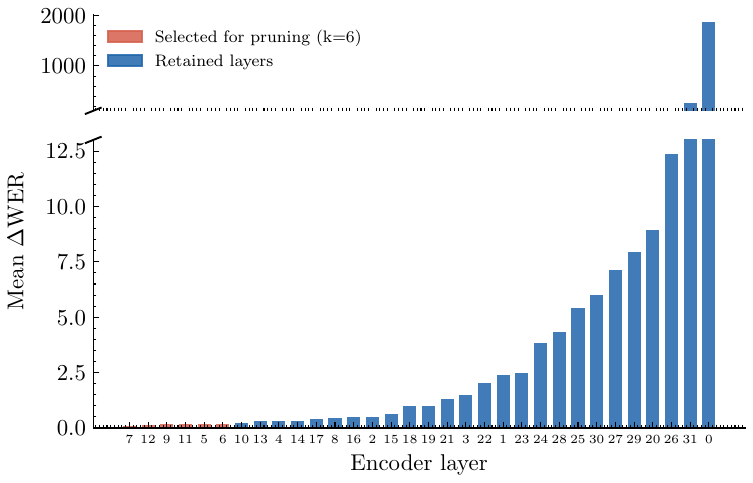}
    \caption{Layer importance scores for all 32 Whisper encoder layers, ranked by mean $\Delta$WER across the selected languages. Orange bars indicate the six layers selected for pruning; blue bars are retained. Layers 0 and 31 are critical layers and cause catastrophic degradation upon removal. The six pruned layers are all located in the start-to-middle of the encoder.}
    \label{fig:layer_ranking}
\end{figure}

Figure \ref{fig:layer_ranking} presents the ranked layer importance scores where we see that the first half of the network is the least sensitive to missing layers with very early layers (0 and 1) being clear outliers, given that these layers transform the signal significantly.

After removing\footnote{Layer removal is done by replacing the model encoder's PyTorch \texttt{nn.ModuleList} with a copy without the selected set of layers.} the $k=6$ least important layers (\texttt{[5, 6, 7, 9, 11, 12]}) we train the resulting zero-shot pruned encoder through knowledge distillation to produce hidden states similar to those of the full encoder through MSE loss. 

We can describe the distillation process as defining the set of layers for removal from the encoder, $\text{Enc}$, as $R=\{L_5, L_6, L_7, L_9, L_{11}, L_{12}\}$ with the resulting pruned encoder being $\text{Enc}_{/R}$. Using unlabeled audio samples $x$, we train the pruned encoder by minimizing MSE loss:

$$
    \mathcal{L}_\text {MSE}(x, \theta_{\text{Enc}_{/R}}) = \frac{1}{n}||  \text{Enc}(x) - \text{Enc}_{/R}(x)||^2_2
$$

We train the pruned encoder (full encoder and decoder are kept frozen) with the AdamW \cite{loshchilov2019decoupledweightdecayregularization} optimizer with a batch size of 8 for 2000 steps on the validation split of the English-only People's Speech \cite{peoples_speech} dataset. Evaluating the mean WER increase for every 500 steps we see convergence during this training configuration which took about half an hour to run on an Nvidia A100 GPU.

\section{Results}

Our primary results are presented in Table \ref{tab:wer_results} where we see that zero-shot pruning incurs a 3.8\% WER increase and distilling closes this gap slightly to an increase of 1.9\% WER.

\begin{table}[tb]
  \centering
  \begin{tabular}{lrrr}
      \toprule
      \textbf{Language} & \textbf{Baseline} & \textbf{Zero-shot} & \textbf{Distilled} \\
      \midrule
      Danish   & 23.9 & 32.1 (+8.2) & 27.3 (+3.4) \\
      English  & 15.4 & 16.6 (+1.2) & 16.1 (+0.7) \\
      German   & 17.1 & 18.3 (+1.2) & 18.1 (+1.0) \\
      French   & 16.3 & 20.7 (+4.4) & 18.7 (+2.5) \\
      \midrule
      \textbf{Mean} & \textbf{18.2} & \textbf{21.9 (+3.8)} & \textbf{20.1 (+1.9)} \\
      \bottomrule
  \end{tabular}
  \caption{WER (\%) on FLEURS language test splits for the full encoder, the zero-shot pruned encoder and after label-free distillation. Parentheses show absolute change in percentage points.}
  \label{tab:wer_results}
\end{table}

Reducing the encoder stack directly decreases memory and storage footprint. Removing six layers eliminates 118M parameters from the encoder and reduces the full model from 1543~MB to 1318~MB in bfloat16 precision (Table~\ref{tab:system_metrics}), a
saving of 225~MB without any change to the decoder or inference code.
\begin{table}[tb]
  \centering
  \begin{tabular}{lrrl}
      \toprule
      \textbf{Metric} & \textbf{Full} & \textbf{Pruned} & \textbf{Change} \\
      \midrule
      Encoder parameters  & 637M   & 519M   & $-$118M  \\
      Model size (bf16)   & 1543MB & 1318MB & $-$225MB   \\
      \bottomrule
  \end{tabular}
  \caption{Memory and storage footprint of the full and pruned model in bfloat16 precision.}
  \label{tab:system_metrics}
\end{table}

To measure inference speedups on consumer hardware we run the model on an Apple M4 Pro running the Transformers \cite{wolf2020transformers} library with Apple's MPS backend. These results are found in Table \ref{tab:mac_benchmark} where we compare the full model to the pruned model on a 60 second out-of-distribution English audio clip. We see that reducing the size of the encoder causes consistent speedups of 1.22-1.24$\times$. It also showcases the time split between the encoder and decoder during end-to-end transcription, highlighting that the encoder is more prominent at 56\% at batch size 8 compared to 36\% at a batch size 1. These speedups and time splits directly lead to greater end-to-end speedups at 1.75$\times$ for batch size 8 versus 1.08$\times$ for batch size 1.

\begin{table}[tb]
  \centering                
  \begin{tabular}{lrrrrr}
      \toprule
      & & \multicolumn{2}{c}{\textbf{Time split}}
        & \multicolumn{2}{c}{\textbf{Speedup}} \\
      \cmidrule(lr){3-4} \cmidrule(lr){5-6}
      \textbf{Model} & \textbf{Batch} & \textbf{Enc} & \textbf{Dec}
          & \textbf{E2E} & \textbf{Enc} \\
      \midrule
      Full (32L)   & 1 & 36\% & 64\% & 1.00$\times$ & 1.00$\times$ \\
      Pruned (26L) & 1 & 32\% & 68\% & 1.08$\times$ & 1.22$\times$ \\
      \midrule
      Full (32L)   & 8 & 56\% & 44\% & 1.50$\times$ & 1.00$\times$ \\
      Pruned (26L) & 8 & 53\% & 47\% & 1.75$\times$ & 1.24$\times$ \\
      \bottomrule
  \end{tabular}
  \caption{Inference benchmark on Apple M4 Pro (\texttt{transformers} + MPS backend), averaged over 5 runs on a 60 second out-of-distribution English audio clip. End-to-end (E2E) speedups are relative to the full model at batch size 1 while encoder (Enc) speedups are relative to the full model at the same batch size.}
  \label{tab:mac_benchmark}
\end{table}

\section{Discussion}

\subsection{Sensitivity to layer selection}
\begin{figure}
    \centering
    \includegraphics[width=\linewidth]{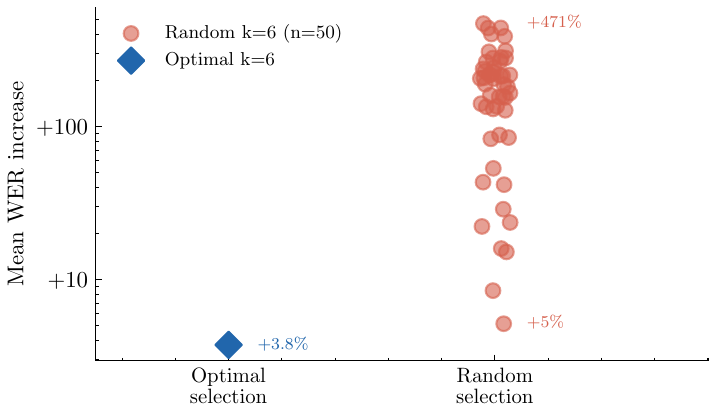}
    \caption{Comparison of the $\Delta$WER-guided optimal layer selection against 50 random selections of six layers drawn from layers 1–15. Random selection spans $+5$ to $+471$ vs. $+3.8$ for the selected set, demonstrating that informed layer selection is essential.}
    \label{fig:random_baseline}
\end{figure}

Is the intricate and rather expensive method of calculating the change in WER across several languages worth it? By randomly sampling $n=50$ combinations of $k=6$ drawn from layers 1-15 we get Figure \ref{fig:random_baseline} which shows that the selection process matters greatly. The redundancy in the early layers is not uniformly distributed and we can retain performance effectively by selecting layers based on the $\Delta$WER importance metric.

\subsection{Removing $>6$ layers}
\label{sec:cliff}
\begin{figure}[tb]
    \centering
    \includegraphics[width=1\linewidth]{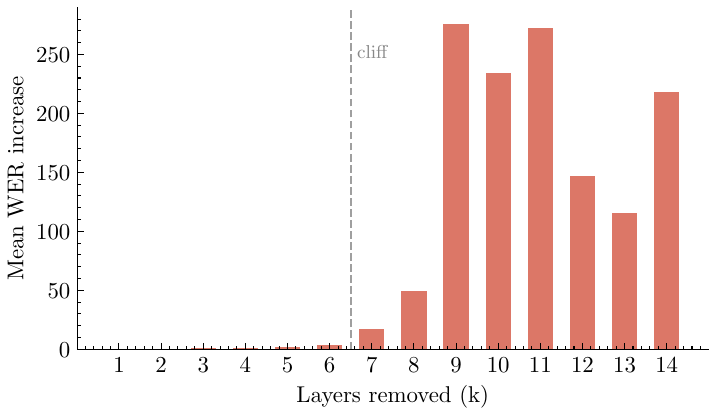}
    \caption{Zero-shot WER degradation (mean across four languages) as a function of the number of layers removed, always taking the $k$ least important layers by $\Delta$WER ranking. Degradation is modest and roughly linear up to $k=6$, after which performance collapses sharply.}
    \label{fig:pruning_sweep}
\end{figure}

\begin{figure}[tb]
    \centering
    \includegraphics[width=1\linewidth]{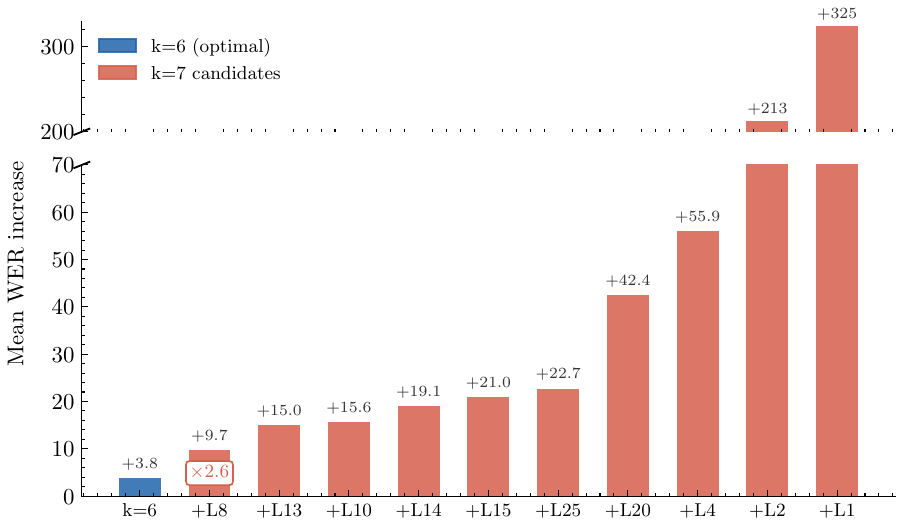}
    \caption{Zero-shot WER degradation when a seventh layer is added to the optimal $k=6$ set, for ten candidate layers. The blue bar shows the $k=6$ reference. Even the most benign seventh layer (L8) causes $2.6\times$ more degradation than $k=6$, confirming a clear pruning boundary.}
    \label{fig:k7_sweep}
\end{figure}

Another aspect regarding sensitivity is to ask why we stop at $k=6$. We investigate this by first measuring the zero-shot mean $\Delta$WER if the process is followed for $k=7$ until $k=14$. The result can be found in Figure \ref{fig:pruning_sweep}. There is a clear cliff-phenomenon wherein performance degradation jumps at $k=7$. Additionally, we look into the sensitivity with regards to selecting the \textit{correct} 7th layer. This is seen in Figure \ref{fig:k7_sweep} where we sweep over a selection of candidates for the 7th removed layer. It shows that even the best 7th layer causes 2.6$\times$ degradation compared to the optimal $k=6$ layer selection. 

Interestingly, the optimal seventh layer to choose is $L_8$, which creates a contiguous 5-layer gap ($L_5$–$L_9$). This could indicate that concentrating the missing processing in one large gap is less damaging than fragmenting it across multiple smaller gaps. That said, the 2.6$\times$ degradation still suggests that this gap exceeds what the encoder can handle.

\subsection{Multilingual recovery through monolingual distillation}

With Danish being the most underrepresented language in the initial training of Whisper it is expected to take the largest hit to performance as seen in Table~\ref{tab:wer_results}. It is, however, also the language that recovers the most despite the data used for distillation being English-only. Notably, distillation recovers all languages despite only training on English audio samples. This would indicate that the label-free distillation through MSE loss on the encoder's hidden states recover acoustic representations rather than language-specific features.

\subsection{Limitations}

While we have shown that multilingual performance is effectively retained, we have limited ourselves to a fairly small selection of languages. A more thorough language-sweep might reveal pitfalls in low-resource languages that are not covered in our work. Similarly, the findings are limited to {\tt whisper-large-v3-turbo}'s encoder stack and results may vary depending on the size of the model.

Distillation was done on an English-only dataset, raising the question of whether multilingual data would improve recovery for those languages that were most negatively affected by layer pruning. Particularly Danish and French, which had the worst performance degradation from the zero-shot pruning but also the greatest recovery from distillation.

\section{Conclusion}

We have demonstrated that layer pruning is entirely feasible for the {\tt whisper-large-v3-turbo} model's encoder. This was done by observing layer redundancy and hypothesizing that several layers could be removed with negligible damage to the model's performance. By using leave-one-layer-out change in Word Error Rate as a metric for layer importance we identify six layers that are the least important and remove them, reducing the encoder stack by 18.5\%. 

We also explore the importance of this metric compared to a random baseline, underlining its significance. Zero-shot pruning raises mean WER from 18.2\% to 21.9\%, which label-free distillation on English-only audio samples recovers to 20.1\% across all languages, suggesting that the MSE training objective realigns acoustic signals rather than language-specific features. 

The pruned encoder is consistently 1.22-1.24$\times$ faster, resulting in $1.75\times$ speedups for larger batch sizes on consumer hardware. Critically, our approach simply produces a shallower network, requiring no changes to inference frameworks, making it a drop-in replacement.

Further research could investigate the cliff-phenomenon discussed in Section \ref{sec:cliff} to better understand why performance suddenly degrades sharply beyond $k=6$. Additionally, future work should investigate whether these findings generalize to more modern ASR models for transcription such as {\tt Cohere Transcribe} \cite{julian_mack_2026}.

\appendix


\bibliographystyle{named}
\bibliography{ijcai26}

\end{document}